# Automated Neural Network Auditing (ANNA) for Energy Market Settlements: A Machine and Reinforcement Learning-Based Approach

**J. R. Rivadeneyra[1], I. Lee[2], F. Nugen[3], S. M. Srinivasan[4]**

**Abstract – Through integration of various segmented energy markets into cohesive Independent System Operators (ISOs) in endeavors such as Energy Imbalance Market (EIMs) of CAISO and MISO energy market settlements have become increasingly complex. The automated process of the transactions in these markets presents an obstacle to the evaluation of the market transactions for accuracy, compliance, and financial integrity. Traditional auditing methods have relied heavily on rule-based manual verification and are prone to inefficiency, human error, and an inability to process large quantities of data produced by automated markets. Conventional limitations support the innovation of an intelligent, scalable anomaly detection of settlement transactions.**

**This research proposes Automated Neural Network Auditing (ANNA), a machine and reinforcement learning framework designed to support settlement auditing through anomaly prioritization. ANNA integrates statistical anomaly detection, K-Nearest Neighbors (KNN), clustering, and Deep Q-Network (DQN) reinforcement learning to identify potentially anomalous settlement transactions from historical market data. The proposed approach is evaluated as an analyst-support framework for improving the prioritization of transactions requiring further audit review.**

**ANNA's performance was evaluated against traditional statistical methods, clustering-based anomaly detection techniques, and supervised and reinforcement learning models. The study draws from three years of historical CAISO settlement data comprising more than 63 million source records across 32 charge codes and multiple counterparties. Following preprocessing and data-quality filtering, approximately 2.9 million labeled transactions constituted the working dataset, from which model-specific experimental subsets were used for training and evaluation. Performance is assessed based on precision, recall, computational efficiency, and false-negative detection. ANNA outperforms traditional anomaly detection methods by identifying hidden relationships in settlement data, while existing approaches, including standard deviation thresholds, K-Means, KNN, and Q-Learning perform poorly across key metrics, highlighting the need for more advanced detection algorithms.**

**The result is a scalable solution for automated auditing of energy market settlements through the integration of machine and reinforcement learning. This work contributes to the field through the support of computationally efficient anomaly prioritization for settlement-auditing workflows.**

## I. INTRODUCTION

### A. Background and Motivation

Increased participation in expanding Independent System Operators (ISOs), Energy Imbalance Markets (EIMs), and Energy Day Ahead Markets (EDAM) has resulted in increased complexity of energy market settlements. ISOs, including the California Independent System Operator (CAISO) and Midcontinent Independent System Operator (MISO), support billions in transactions and settlements at ever-increasing levels of granularity, requiring verification, transparency, and compliance [1]. The sheer volume of transactions of these systems represents a substantial challenge for auditing and anomaly detection [2]. Large-scale anomaly detection challenges have also been documented in broader financial and operational datasets [35]. Traditional auditing methods, such as rule-based auditing with manual verification for discrepancy identification, are infeasible due to the scale and complexity of modern energy markets [3].

[1] Pennsylvania State University Department of Engineering, University Park, PA 16802
[2] UC Berkeley Department of Electrical Engineering and Computer Science, Berkeley, CA 94720
[3] UC Berkeley School of Information, Berkeley, CA 94720
[4] Pennsylvania State University Department of Engineering, Great Valley, Malvern, PA 19355-1443

Existing fixed rule-based and statistical anomaly energy settlement auditing frameworks lack the adaptability needed for modern energy markets, resulting in numerous missed anomalies [4]. The inability of a rule-based auditing framework to adapt to changing market conditions is unsuitable for the evolving trading market with dynamic price models, novel charge codes, shifting regulatory policies, and hidden relationships traditional methods cannot hope to capture [5]. Compounded with the extensive labor and expertise manual audits require, traditional methods are simply impractical for the large-scale operations of modern markets [6]. Current auditing inefficiencies lead to financial losses, compliance risks, inaccurate settlements, and lost cost recovery opportunities, impacting market participants and regulators [7]. From an innovation economics perspective, AI-driven auditing represents a technological disruption of established institutional practices in regulated energy markets, where the adoption of intelligent systems can reduce transaction costs and improve market integrity.

**B. Problem Statement**

The principal issue in auditing modern energy market settlements is the inefficiency of traditional methods when applied to large-scale datasets with transactions occurring at higher frequencies than traditional transactions. Traditional fixed-threshold statistical models' reliance on predefined confidence intervals fails to detect subtle context-dependent anomalies [8]. Lack of subtlety and scalability make the real-time processing of large volumes of data impossible [9].

Several key challenges with existing auditing frameworks:

1. **Volume of Transactions:** ISOs process millions of settlement records at different frequencies daily, making manual audits impractical [10].
2. **Complexity of Charge Codes:** Different resource and settlement types involve variable price structures, interval pricing, and time-based adjustments, making it difficult to identify inconsistencies with traditional methods [11].
3. **Limited Adaptability of Existing Models:** Existing models fail to adjust dynamically to resource and market changes, new charge codes, and evolving pricing structures [12].
4. **High False Positive and Omission Rates:** Rule-based detection models generate excessive false positives or fail to identify anomalies, leading to an increased burden on financial analysts as well as missed opportunities for cost recovery [13].

From a systems perspective, modern energy settlement markets represent highly dynamic and nonstationary environments where transaction behaviors evolve continuously in response to pricing volatility, regulatory adjustments, resource participation changes, and fluctuating demand conditions. Traditional auditing methods are generally designed around static assumptions and predefined rule structures, limiting their ability to adapt as settlement behaviors shift over time. This creates a growing need for auditing frameworks capable of learning from sequential transaction patterns while adjusting to evolving market conditions without requiring constant manual recalibration.

Reinforcement learning provides a theoretical foundation well suited to these conditions because it enables decision policies to evolve through iterative interaction with changing environments. Unlike conventional supervised approaches that rely heavily on fixed historical labels, reinforcement learning frameworks evaluate actions through accumulated reward feedback, allowing the model to adapt to complex temporal relationships and emerging settlement behaviors. Within energy markets, where transaction irregularities often develop through sequences of interconnected operational events rather than isolated deviations, sequential learning methods offer meaningful advantages over static threshold-based approaches.

**C. Research Objectives**

The objective of this research is the development of an Automated Neural Network Auditing (ANNA), an AI-based auditing system for enhanced efficiency, accuracy, and scalability of power market settlements. ANNA integrates multiple layers of anomaly detection for understanding and incremental improvement, starting with statistical anomaly detection, followed by machine learning (ML) and reinforcement learning (RL) to perform adaptive auditing. Through the application of neural networks to analyze historical settlements data, identify patterns, and predict potential discrepancies before they impact financial statements, the ANNA deviates from conventional auditing systems [14].

Key research objectives include:

- The detection and prioritization of settlement anomalies using machine learning algorithms.
- Support for settlement review and compliance-monitoring workflows.
- Reduction in financial risks through early error detection.
- Improved anomaly detection and accuracy with Deep Q-Network (DQN) reinforcement learning.
- Reduction and minimization of false positives through the integration and application of K-Nearest Neighbors (KNN) clustering and historic benchmarking.

### D. Contributions

This research makes several significant contributions to the field of automated financial auditing in power markets. First, the research introduces an AI-based framework that extends traditional rule-based methods with a scalable intelligent auditing system. Second, the research integrates and applies multiple machine learning techniques, including neural networks, clustering, and reinforcement learning, with traditional statistical techniques to improve the precision and adaptability of anomaly detection. Third, the research draws upon three years of CAISO settlement data, with the complete source dataset comprising more than 63 million transaction records across 32 charge codes for Western Energy Imbalance Market (WEIM) participants. Following preprocessing and data-quality filtering, approximately 2.9 million labeled transactions constituted the working dataset, from which model-specific experimental subsets were used for training and evaluation to maintain computational tractability while preserving representative operational and settlement characteristics [15]. Fourth, the methodology proposed optimizes computation efficiency by supporting human auditors as a primary filter for high-risk transactions. Finally, the ANNA provides a scalable architecture applicable across multiple ISOs, ensuring a blend of accuracy with adaptability in different power markets [16].

The contribution of this research lies primarily in the integration and operationalization of reinforcement learning techniques within large-scale settlement auditing workflows rather than the development of a fundamentally new learning algorithm. The framework emphasizes scalable anomaly prioritization, adaptive transaction analysis, and practical deployment considerations within highly dynamic market environments. Rather than treating settlement anomalies as isolated statistical outliers, the proposed framework approaches anomaly identification as an evolving sequential decision problem influenced by operational context, temporal market behavior, and historical settlement interaction patterns. This perspective aligns more closely with the realities of modern ISO market operations, where irregular settlement behavior often develops progressively across multiple market intervals and transaction states.

### E. Structure of the Paper

The remainder of this paper is structured as follows: Section II reviews related work such as existing auditing methods, machine learning techniques for anomaly detection, and the application of reinforcement learning in financial markets. Section III outlines the proposed ANNA method, including data preprocessing, applicable feature engineering, and training procedures for neural networks. Section IV details the experimental setup, data characteristics, and metrics for evaluating the performance of the ANNA. Section V discusses and compares the results between traditional statistical methods, cluster-based anomaly detection, and reinforcement learning. Section VI specifies challenges, limitations, and future improvements in AI-based settlements auditing based on the results of this research. Section VII summarizes the key findings, implications, and recommendations of this research for application in future power market anomaly detection.

## II. RELATED WORK

### A. Existing Auditing Techniques in Energy Markets

Traditional auditing techniques for power markets settlements rely on rule-based validation, statistical threshold-based anomaly detection, and manual inspection for final validation. Comparing settlement data to predefined rules established by Independent System Operators (ISOs) and regulatory authorities provides an inherently rigid, structured anomaly detection incapable of adapting to evolving market conditions [17]. Statistical threshold-based methods, such as Z-score anomaly detection and confidence interval-based filtering, struggle to differentiate natural market fluctuation from genuine settlement discrepancies [18]. Recent work highlights similar challenges in real-time anomaly detection across complex energy systems [19, 20].

Historically, manual auditing played a significant role in power market settlements; however, the rapid increase in transaction volume and charge-code complexity associated with the integration of technologies such as renewable energy resources has rendered manual verification increasingly impractical. Studies show manual auditing in high-volume power markets results in inefficiencies, increased costs, and missed anomalies [5]. Addressing these challenges has led researchers to explore automated anomaly detection mechanisms using traditional statistical methods and machine learning-based approaches.

### B. Machine Learning for Anomaly Detection

The application of machine learning (ML) in financial and power market anomaly detection continues to gain interest in recent years. ML models are capable of automatically identifying patterns, detecting outliers, and improving anomaly detection accuracy through continuous training from historical transaction data [4]. Recent AI enabled anomaly detection frameworks further demonstrate the importance of ML in high-volume energy systems [21-23]. Several ML approaches proposed for financial settlement verification include unsupervised clustering techniques, classification models, and deep learning architectures.

A common approach for unsupervised anomaly detection utilizes algorithms such as K-means clustering, Isolation Forests, and One-Class Support Vector Machines (SVMs) to identify transactions deviating from historical norms [2]. These models are suitable for high-volume financial settlement by effectively grouping similar transactions to highlight anomalies without labeled data. However, their effectiveness is limited to low-dimensional datasets and simple market behaviors, resulting in false positives and low adaptability to novel market conditions common with industry-wide disruptions [13]. Additional studies highlight the role of big-data ML techniques in electricity consumption anomaly detection [24].

Supervised learning techniques train models on historical transaction labels such as “normal” or “anomalous”. Studies show neural networks, gradient-boosting models, and ensemble learning methods enhance the accuracy of anomaly detection in power settlements [9]. However, supervised learning models’ need for datasets with a high percentage of accurate labels is limiting and infeasible in many instances of power market applications.

Advancements in deep learning-based anomaly detection have introduced novel techniques such as Autoencoders and Long Short-Term Memory (LSTM) networks to capture complex temporal dependencies in settlements [1]. These methods show promise in the subtle identification of context-dependent anomalies common in financial records. However, deep learning models are challenging to deploy in a real-time audition environment due to their significant computation requirements, large-scale training datasets [3], and data drift.

### C. Reinforcement Learning in Auditing and Anomaly Detection

Exploration in Reinforcement Learning (RL) has proved promising as an alternative to traditional anomaly detection methods in financial and power market auditing. Unlike supervised or unsupervised learning approaches, reinforcement learning allows adaptive decision-making through dynamic adjustment of anomaly detection thresholds based on market behaviors [12].

Deep Q-Network (DQN) learning is a widely adopted reinforcement technique for anomaly detection using Q-learning combined with deep neural networks to optimize decision-making in dynamic environments [7]. The application of DQN in fraud detection, credit risk assessment, and financial market anomaly detection has delivered promising results [13]. Recent RL-based anomaly detection research reinforces the suitability of RL for dynamic settlement environments [25–28]. Additional systematic reviews further highlight the growing role of deep reinforcement learning in anomaly detection across diverse operational domains [29]. In the power market settlements context, reinforcement learning can continuously update its anomaly detection strategy based on new transaction data, resulting in a highly adaptable approach for real-time auditing.

Existing RL studies for anomaly detection have demonstrated RL agents' trainability for settlement transaction binary classification as "normal" or "anomalous" based on learned patterns [9]. Combined with reward-based optimization, RL enables the model to prioritize high-risk anomalies while minimizing false positives. Limited research on applying RL to large-scale power settlements highlights the need for further investigation [1].

### D. Gaps in Existing Research and Motivation for ANNA

Research gaps persisted in the power market settlement context despite the advancements made in ML and RL for anomaly detection. Existing studies focus on financial fraud detection with limited applications for ISO settlement charge codes and market structures [15]. Similarly, traditional ML-based approaches have not had sufficient adaptability to handle evolving charge codes, resource mixes, market volatility and new regulatory requirements [16]. Reinforcement learning applications within energy systems have primarily focused on optimization problems such as load balancing, dispatch scheduling, and dynamic pricing. Comparatively limited research has explored reinforcement learning within settlement auditing environments, where anomaly detection involves high-dimensional transactional behavior, temporal dependency structures, and continuously evolving market conditions. This distinction is important because settlement anomalies frequently emerge through combinations of operational, financial, and timing-related interactions that are not easily captured through static classification methods alone. Recent studies on anomaly detection in power markets underscore the need for scalable RL-based auditing frameworks [30].

The proposed model, Automated Neural Network Auditing (ANNA), addresses these gaps by introducing a hybrid anomaly detection framework integrating statistical benchmarking, unsupervised clustering, and reinforcement learning. Through the combination of multiple detection methodologies, ANNA is designed to support the identification and prioritization of potentially anomalous settlement transactions for further audit review. Additionally, ANNA utilizes reinforcement learning to incorporate asymmetric operational consequences through reward-based decision-making, allowing different penalties to be assigned to false-positive and false-negative classifications. Future extensions may incorporate online learning and richer state representations to improve adaptation to evolving market conditions and data distributions[17].

Presently, RL models for anomaly detection have limited application to real-world energy settlement due to being evaluated on synthetic datasets or small-scale financials. This research aims to bridge this gap by drawing on three years of CAISO settlement data comprising more than 63 million source charge-code transactions across 5- and 15-minute market intervals and multiple resources. Following preprocessing, approximately 2.9 million labeled transactions constituted the working dataset, from which model-specific experimental subsets were used for training and evaluation. This large-scale evaluation will provide additional insights into the scalability and real-world feasibility of AI-driven anomaly detection in energy settlements [18]. Complementary research in local energy markets also demonstrates the growing role of ML and RL in decentralized settlement processes [31].

The proposed ANNA model addresses these research gaps by providing an automated framework for settlement anomaly identification and prioritization.

## III. PROPOSED METHODOLOGY

This section outlines the methodological framework of ANNA, including the data preprocessing procedures, feature engineering strategies, and the machine learning model architecture and training protocols applied to neural network development and reinforcement learning-based anomaly detection.

### A. Dataset Summary

The source data for the experimental evaluation of the proposed Automated Neural Network Auditing (ANNA) framework comprised more than 63 million real-world power market settlement transactions obtained from the California Independent System Operator (CAISO). Following preprocessing and data-quality filtering, approximately 2.9 million labeled transactions constituted the working dataset, from which model-specific experimental subsets were used for training and evaluation. The dataset spans the three-year period from January 2021 through December 2023 and includes settlement activity across multiple market intervals, resource types, and charge code structures. This time frame incorporates a variety of resource changes, market conditions, seasonal trends, and policy adjustments, representing a robust foundation for model training and anomaly detection evaluation [13].

Each transaction includes metadata defining the Position Name, Settlement Types, Trade Date, Invoice Date, Hour Ending, Charge / Determinant Name, corresponding Amount / Value, Hour Ending, Invoice Date, and ISO / Shadow delineations.

Key data features of the over 32 unique charge codes representing different financial and operational components of the ISO settlement process include the prominent operation charge codes of:

- **Real-Time Locational Marginal Pricing (LMP):** Captures granular pricing variations across 5-minute intervals to reflect market supply and demand conditions at specific nodes.
- **Interval Energy Imbalance Settlements:** Accounts for discrepancies between forecasted and actual energy consumption or generation within defined intervals.
- **Daily Market Forecast and Recalculation Charges:** Include initial settlement estimations and subsequent recalculations based on updated inputs or corrections.
- **Market Rerun Adjustments:** Capture retroactive settlement corrections stemming from administrative updates, regulatory compliance, or pricing rule changes.

Due to the scale of the full settlement dataset, model training and evaluation were performed on filtered and grouped subsets partitioned by Position Name, Settlement Type, and charge code characteristics. This localized training strategy enabled computationally feasible experimentation while preserving the broader statistical and operational properties represented throughout the complete 63 million record dataset. The full dataset remained available for preprocessing, aggregation, benchmarking, and comparative analysis throughout the study.

Each respective charge code exhibits different temporal and volumetric characteristics for the model to learn diverse transaction behaviors across a baseline of settlements and corrective adjustments. Among these, Real-Time LMP and Interval Energy Imbalance settlements tend to exhibit greater anomaly frequency due to their sensitivity to short-term price volatility and forecast deviations, while rerun adjustments are more prone to systematic errors arising from retroactive corrections.

Although localized subsets were used during individual training cycles, the broader dataset structure preserved temporal variability, charge code diversity, and settlement behaviors representative of large-scale ISO market operations. Grouped processing strategies were necessary to support memory-efficient model training and iterative reinforcement learning procedures while maintaining exposure to the broader statistical characteristics present throughout the complete settlement dataset.

**B. Data Preprocessing**

Prior to training and evaluation, the data was subjected to systematic preprocessing to ensure consistency, accuracy, and compatibility with machine learning models.

First, all nonnumeric values in the Amount column were converted to floating-point format using robust typecasting. Entries with invalid or missing values were dropped. Transactions identified as "Shadow" in the ISO / Shadow column were excluded to target the analysis on finalized market data. Datetime objects were utilized to extract time-based features, including month, weekday, and hour.

To enable localized anomaly detection tailored to individual units and market mechanisms, the dataset was filtered by Settlement Types and Position Name. The data was then grouped by these contextual identifiers, supporting the learning model's identification and adoption of transaction patterns of each market participant, significantly improving interpretability and anomaly resolution.

Anomaly labels for model evaluation were derived using a composite rule-based methodology combining statistical thresholds (values exceeding ±2 standard deviations from localized rolling means) with domain-informed heuristics associated with known settlement correction behaviors, rerun adjustments, and irregular transaction patterns observed during historical audit review. These labels were intended to provide an approximate operational benchmark for anomaly identification rather than definitive ground-truth classifications.

To improve confidence in the generated labels, representative flagged transactions were manually reviewed against available settlement documentation and historical correction records. The review focused primarily on high-value deviations, recurring adjustment patterns, and transactions exhibiting significant divergence from expected settlement behavior. However, the manual review was conducted as an operational plausibility check rather than as a formally designed independent validation study; therefore, a predefined validation sample size, standardized anomaly-confirmation protocol, and formal inter-rater agreement assessment were not established. Due to the scale of the dataset exceeding 63 million records, comprehensive manual validation was not operationally feasible. This limitation restricts the extent to which the reviewed transactions can be considered independently verified ground truth.

While the labeling methodology provided a practical foundation for large-scale anomaly detection evaluation, the potential for label uncertainty, heuristic bias, and partial circularity remains a limitation of the study. Because the proxy labels were derived partly from statistical deviation criteria and domain-informed heuristics, models that identify patterns similar to those embedded in the labeling rules may receive more favorable evaluation results even when the corresponding transactions have not been independently confirmed as settlement errors. Consequently, the reported performance metrics should be interpreted as measures of agreement with the operational proxy-labeling framework rather than as performance against independently verified ground truth. This potential label-induced bias should therefore be considered when interpreting comparative model performance.

The anomaly labeling process was designed to emulate practical auditing workflows commonly used in large-scale settlement review environments, where exhaustive manual verification is typically infeasible due to transaction volume and operational constraints. Consequently, the resulting labels should be interpreted as operational indicators intended to support anomaly prioritization rather than absolute determinations of settlement error. The manual review process was intended to support operational plausibility assessment rather than provide a statistically representative, independently validated ground-truth dataset. Additional details concerning proprietary settlement review procedures and transaction-level validation are subject to applicable data-use and confidentiality restrictions. Consequently, the reported metrics reflect agreement with the proxy-labeling methodology and should not be interpreted as independently verified settlement-error detection rates.

The complete dataset comprised 63,693,428 settlement records collected from 29 source files spanning January 2021 through January 2024. Following preprocessing and data-quality filtering, 2.9 million labeled transactions were loaded for analysis, representing 447 settlement-position groups, 116 unique position names, 11 charge codes, and 5 settlement types across 3 counterparties. The reduced working dataset was selected to accommodate available system memory while preserving the temporal ordering and representative characteristics of the original 63.7 million-record dataset. The operational labeling process identified 144,707 anomalous transactions (4.99%).

## Experimental Dataset Characteristics

| Dataset Component | Value |
|---|---|
| CSV source files | 29 |
| Total source records | 63,693,428 |
| Records loaded after preprocessing | 2,900,000 |
| Usable labeled records | 2,900,000 |
| Normal transactions | 2,755,293 |
| Anomalous transactions | 144,707 |
| Anomaly prevalence | 4.99% |
| Unique settlement types | 5 |
| Unique position names | 116 |
| Unique settlement-position groups | 447 |
| Unique charge codes | 11 |
| Unique counterparties | 3 |
| Earliest transaction date | January 1, 2021 |
| Latest transaction date | January 5, 2024 |
| Random seed | 42 |

*Table 1 - Experimental Dataset Characteristics*

To clarify the experimental sample sizes and data-partitioning procedures used across methods, the model-specific configurations are summarized in Table 2. The reinforcement learning models were trained using identical partitions consisting of 6,000 training, 1,500 validation, and 2,500 testing records. Where transaction timestamps were available, the partitions were created chronologically to better approximate real-world deployment and reduce the potential for temporal data leakage. Statistical and unsupervised methods operated on the complete or sampled datasets appropriate to their respective implementations.

## Experimental Data Partitioning by Model

| Method | Maximum Rows | Rows Used | Data Partitioning |
|---|---|---|---|
| Statistical (±2 SD) | All available | 2,900,000 | Full dataset; no train/test partition |
| K-Nearest Neighbors | 10,000 | 10,000 | Full sampled dataset |
| K-Means Clustering | 10,000 | 10,000 | Full sampled dataset |
| Q-Learning | 10,000 | 10,000 | 6,000 train; 1,500 val.; 2,500 test |
| ANNA (DQN) | 10,000 | 10,000 | 6,000 train; 1,500 val.; 2,500 test (chronological when available) |

*Table 2 - Experimental Data Partitioning by Model*

### C. Feature Engineering

Raw settlement records were transformed into structured, learnable representations:

- Temporal indices: Derived from parsed Trade Date and Invoice Date fields, including Month, Day of Week, and Hour Ending.
- Categorical encodings: Like Charge / Determinant Name and Settlement Type, encoded using one-hot or label encodings depending on model requirements.
- Benchmark statistics: Local rolling mean and standard deviation were calculated for each charge code within a group to normalize transactional behavior over time.

Feature development and anomaly characterization were informed in part by historical settlement review practices and operational audit behaviors commonly used in market reconciliation procedures. Incorporating domain-informed contextual features improved the framework's ability to distinguish localized settlement irregularities from routine market volatility.

**D. Neural Network Architecture**

The ANNA detection engine incorporates a hybrid modeling architecture combining statistical benchmarking, unsupervised clustering, and reinforcement learning. The principal learning agent is the Deep Q-Network (DQN), learning to identify anomalous settlements by optimizing a given reward function tied to transaction deviations and model uncertainty.

Q-learning operates within a simulated Gymnasium-based environment, where each transaction index represents a state and a binary classification where normal or anomalous represents the action. The environment returns rewards based on the action's alignment with statistically deviant behavior (e.g., transaction is outside a learned confidence interval). The reward function applies an asymmetric penalty structure, where false negatives (missed anomalies) are penalized more heavily than false positives to reflect the higher cost of undetected settlement errors in financial auditing. It is important to note that Q-Learning's tabular structure, while effective for small discrete state spaces, is susceptible to memorization of exact training patterns when the state-action space is limited, which may result in overfitting rather than generalization.

The decision-making process is improved through iterative updates of the Q-table using the Bellman equation:

$$Q(s,a) \leftarrow Q(s,a) + \alpha \left[ r + \gamma \cdot \max_{a'} Q(s', a') - Q(s,a) \right]$$

Where $\alpha$ is the learning rate, $\gamma$ is the discount factor, $r$ is the reward, $s$ and $a$ are the current state and action, and $s'$ and $a'$ are the next state and subsequent action. Anomalies are identified when the model encounters high temporal-difference errors, indicating a transaction does not conform to the expected behavioral policy. ANNA targets settlement calculation errors and billing discrepancies rather than operationally legitimate price deviations caused by extreme weather events or sudden technical failures. The framework relies on historical benchmarking and contextual charge code features to differentiate between these categories, though further refinement of this distinction remains a direction for future work.

To replace the static lookup table, the DQN trains a neural network to approximate Q-values for each possible action, given the current state as input. The loss function minimized during training is based on the squared difference of the predicted Q-value and the target computed from the Bellman equation:

$$\text{Loss} = \left[ Q(s,a) - \left( r + \gamma \cdot \max_{a'} Q(s', a') \right) \right]^2$$

Experienced replay stabilizes learning by storing previous state-action-reward sequences to be sampled randomly to train the network. A separate target network is periodically updated to reduce instability during training.

The DQN implementation utilized a feedforward neural network architecture consisting of a one-dimensional standardized settlement Amount input, followed by one fully connected hidden layer containing 64 neurons with a Rectified Linear Unit (ReLU) activation function. The output layer consisted of two linear units representing the Q-values associated with the normal and anomalous classification actions. The network was trained using the Adam optimizer with a learning rate of 0.001 and mean squared error (MSE) loss. The reinforcement learning configuration used a discount factor ($\gamma$) of 0.90, a mini-batch size of 64, and an experience replay memory capacity of 10,000 transitions. Training was performed over 100 episodes using randomized mini-batches sampled from replay memory. Exploration followed an epsilon-greedy policy initialized at 1.0 and progressively decayed by a factor of 0.995 to a minimum value of 0.01. The reward function assigned +10 for correctly identifying an anomalous transaction, +1 for correctly identifying a normal transaction, −5 for incorrectly flagging a normal transaction as anomalous, and −10 for failing to identify an anomalous transaction.

To improve the reproducibility of the proposed Automated Neural Network Auditing (ANNA) framework, the DQN architecture is described in terms of its input representation, neural network topology, reinforcement learning configuration, exploration strategy, reward formulation, experience replay mechanism, and principal training hyperparameters implemented in the experimental codebase. Specifically, the DQN configuration used to generate the reported results consisted of a 1–64–2 network architecture, comprising one standardized settlement Amount input, one 64-neuron ReLU hidden layer, and two linear output units corresponding to the normal and anomalous actions. This architectural description complements the mathematical formulation presented in this section and provides a concise representation of the implemented DQN learning pipeline.

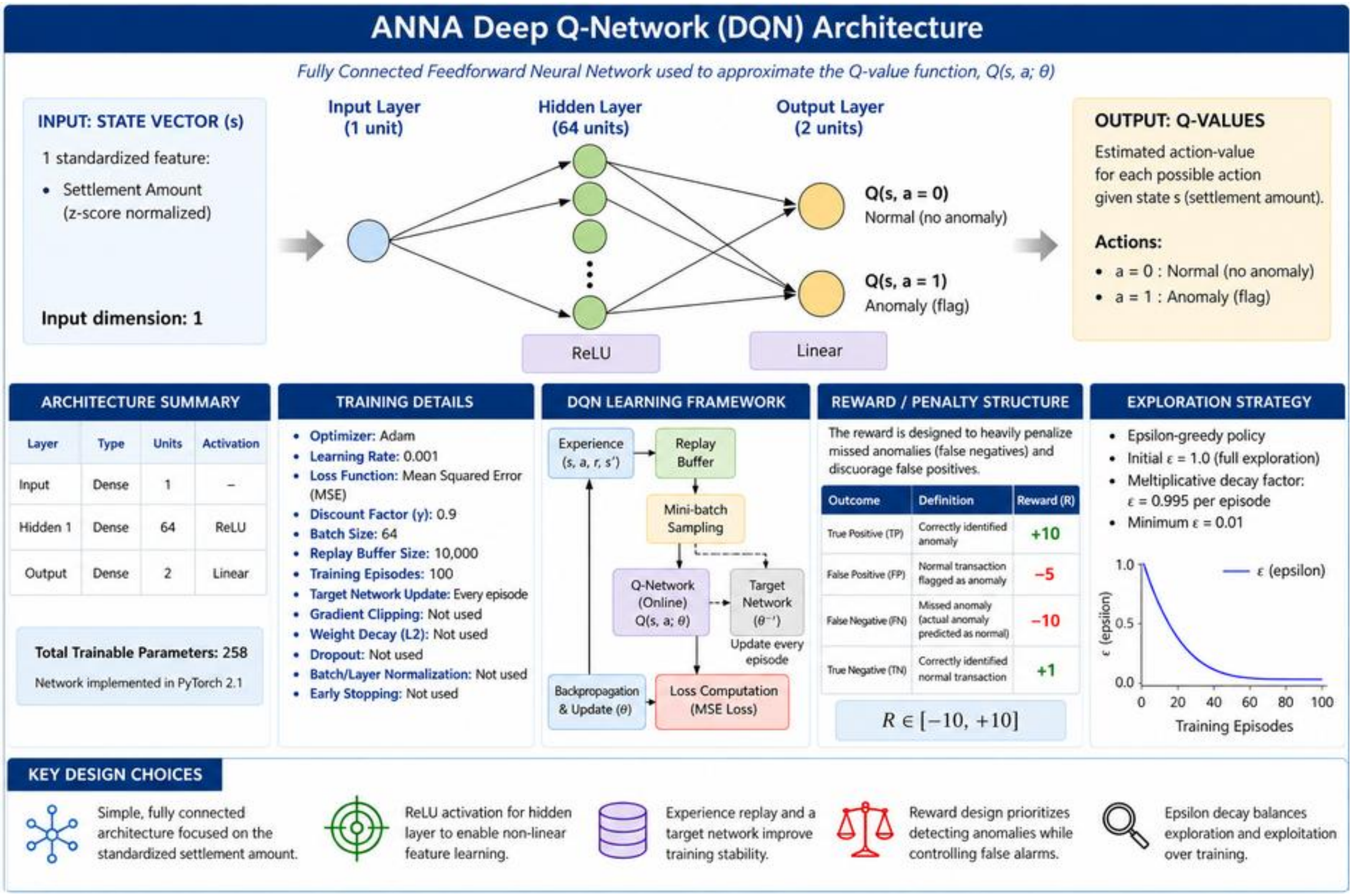

The ANNA agent receives a one-dimensional state representation consisting of the standardized settlement Amount for each transaction and estimates the action-value function $Q(s, a, \theta)$ for two actions corresponding to normal and anomalous classifications. The DQN consists of a single fully connected hidden layer containing 64 neurons with a rectified linear unit (ReLU) activation function, followed by a two-unit linear output layer that produces the estimated Q-values for the two actions. The network contains 258 trainable parameters. Training uses experience replay with a replay-buffer capacity of 10,000 transitions, a target network updated after each training episode, and an $\epsilon$-greedy exploration strategy. The network is optimized using the Adam optimizer with a learning rate of 0.001 and mean squared error (MSE) loss. The discount factor $\gamma$ is set to 0.9, the mini-batch size is 64, and training is performed for 100 episodes. Exploration begins at $\epsilon = 1.0$, decreases multiplicatively by a factor of 0.995 per episode, and is bounded by a minimum value of 0.01. These mechanisms provide the exploration, replay, and target-network components used in the implemented DQN training procedure.

Although the anomaly-detection task can also be formulated as a supervised binary classification problem using the proxy labels generated in this study, the DQN framework enables asymmetric operational costs to be incorporated directly through the reward function. In the implemented reward structure, correctly identified anomalies receive a reward of $+10$, correctly identified normal transactions receive $+1$, false-positive anomaly flags receive $-5$, and false negatives receive $-10$. This asymmetric reward structure places greater emphasis on missed anomalies while also penalizing false alarms. A supervised multilayer perceptron (MLP) is therefore included as a comparative baseline to evaluate whether the DQN formulation provides measurable anomaly-discrimination benefit beyond conventional supervised neural classification under the evaluated experimental configuration.

In the current implementation, each settlement transaction is represented as an environment state consisting of the standardized settlement Amount, and the agent selects one of two actions: classify the transaction as normal or anomalous. Transactions are processed independently, and the present state representation does not explicitly encode long-term temporal dependencies among settlement records. Consequently, the current DQN should not be interpreted as demonstrating that sequential dependencies between transactions have been learned. Rather, the reinforcement-learning formulation provides a configurable reward-based decision framework that can be extended in future work to incorporate richer state representations, temporal dependencies, concept drift, adaptive thresholds, online learning, and dynamically changing reward structures.

Given this independent-transaction formulation, the present DQN shares characteristics with cost-sensitive binary classification. The reinforcement-learning formulation is used primarily to incorporate asymmetric operational consequences through the reward function, allowing different penalties to be assigned to false-positive and false-negative decisions. To evaluate whether this reward-based formulation provides measurable benefit beyond conventional supervised neural classification, a supervised multilayer perceptron (MLP) was implemented using the same feature representation, proxy labels, preprocessing procedures, and data partitions as the DQN. Under the evaluated experimental configuration, the DQN achieved higher anomaly-discrimination performance than the MLP across the reported precision, recall, F1, ROC-AUC, and PR-AUC metrics. These results are specific to the present dataset, proxy-labeling procedure, feature representation, and experimental configuration and are not intended to establish a general superiority of reinforcement learning over supervised classification.

Hyperparameter selection was guided through iterative experimental evaluation emphasizing training stability, convergence consistency, and computational feasibility across grouped settlement transaction subsets. The final configuration used one 64-unit hidden layer, a learning rate of 0.001, a discount factor of 0.9, a mini-batch size of 64, a replay-buffer capacity of 10,000 transitions, and 100 training episodes. No dropout, weight decay, batch/layer normalization, gradient clipping, or early stopping was applied in the reported configuration.

**Table 3. DQN Architecture and Training Hyperparameters**

| Parameter | Value | Parameter | Value |
|---|---|---|---|
| Input features | 1 (standardized settlement Amount) | Batch size | 64 |
| Hidden layers | 1 | Replay buffer size | 10,000 |
| Hidden units | 64 | Training episodes | 100 |
| Hidden activation | ReLU | Target network update | Every episode |
| Output units | 2 | Exploration policy | Epsilon-greedy |
| Output activation | Linear | Initial ε ($\varepsilon_{start}$) | 1.0 |
| Network architecture | 1 – 64 – 2 | ε decay rate | 0.995 per episode (multiplicative decay) |
| Total trainable parameters | 258 | Minimum ε ($\varepsilon_{min}$) | 0.01 |
| Optimizer | Adam | Gradient clipping | Not used |
| Learning rate | 0.001 | Weight decay (L2) | Not used |
| Loss function | Mean Squared Error (MSE) | Dropout | Not used |
| Discount factor (γ) | 0.9 | Batch/Layer normalization | Not used |
| | | Early stopping | Not used |

*Table 3: DQN Training Configuration Parameters*

With the incorporation of DQN into the framework, ANNA provides a reward-based anomaly classification mechanism in which asymmetric operational costs can be incorporated directly into the learning objective. Under the evaluated configuration, the DQN learns a decision policy from the standardized settlement Amount and the specified reward structure. The current implementation does not explicitly model temporal dependencies or demonstrate online adaptation to changing data distributions. Future work will therefore investigate richer state representations, recurrent and attention-based architectures, adaptive replay strategies, and online learning mechanisms to improve temporal pattern recognition and adaptation to evolving settlement conditions.

All reported DQN experimental results were generated using the documented 1–64–2 network configuration, with standardized settlement Amount as the sole input feature. Other contextual variables described in the preprocessing and feature-engineering sections were used for data organization and analysis and should not be interpreted as additional inputs to the reported DQN architecture.

### E. Computational Environment

All experiments were conducted on a Windows workstation running Python 3.10.7 and PyTorch 2.5.1. The system used an Intel Core i7-1165G7 processor with four physical cores (eight logical processors) and 15.65 GB of system memory. The installed PyTorch build was CPU-only (v2.5.1+cpu); therefore, CUDA acceleration was unavailable and no NVIDIA GPU was used during model training or inference. Wall-clock runtime for each anomaly detection method was measured automatically by the evaluation pipeline. Reported execution times include model initialization, data preprocessing performed within each model, training, validation-based threshold selection, and inference, but exclude dataset loading, visualization generation, and export of results.

### F. Training Procedure

Model training was across individual groups defined by Position Name and Settlement Type to allow localized learning. Multiple iteration episodes were performed for each data group, with exploration gradually decaying through an epsilon-greedy policy. The RL training procedure uses reward-based optimization to learn anomaly-classification decisions under the specified experimental configuration. To reduce overfitting and improve generalization, training and evaluation partitions were separated chronologically wherever possible, preventing future settlement records from influencing earlier policy learning stages. Validation subsets were additionally used to evaluate policy stability and threshold sensitivity prior to final testing.

Precision, recall, and delta-based metrics are used to evaluate performance feedback supporting the comparison of model-detected outliers to known statistical outliers. Additionally, predicted values (*Forecasted Amount*) for each identified anomaly are retained to support post-audit reconciliation and review. The term "global statistics" as referenced in this work refers to the aggregate mean and standard deviation metrics computed across each charge code group; these statistics form the basis of the Statistical (±2 SD) thresholding method used as the baseline comparator in the evaluation.

Fig. 1 below presents the DQN (Time) model's training performance over 100 episodes. The left subplot illustrates model accuracy converging rapidly and remaining stable, while the right subplot shows a consistent decline in total loss, indicating effective learning:

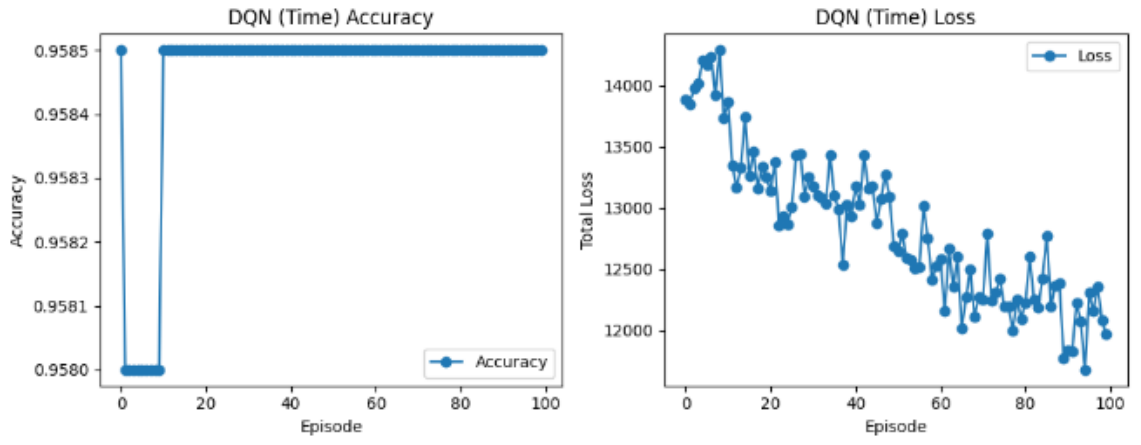

*Fig. 1: Model Training Performance.*

The ANNA system implemented utilizes Python with key libraries including NumPy, Pandas, Scikit-learn, PyGAM, and Gymnasium. Modular implementation supports scalability and portability across different energy markets and audit environments.

**G. Validation and Generalization Strategy**

To evaluate the robustness and generalization capability of the proposed framework, the dataset was partitioned using temporally aware training, validation, and testing procedures. Settlement records were separated chronologically to reduce the possibility of future transaction information influencing earlier model training stages. This approach better reflects operational deployment conditions where anomaly detection models must evaluate newly arriving market transactions without prior exposure to future settlement behavior.

For reinforcement learning experiments, grouped transaction subsets were divided into training and evaluation partitions using sequential holdout windows. Validation datasets were utilized to calibrate anomaly thresholds, evaluate policy stability, and reduce the likelihood of overfitting to repetitive settlement behaviors. Test partitions were retained exclusively for final performance evaluation.

Due to the highly imbalanced nature of settlement anomaly detection, performance assessment emphasized precision, recall, F1-score, and AUC-ROC rather than accuracy alone. This prevented majority-class settlement records from disproportionately influencing performance interpretation.

Additional precautions were taken to reduce memorization effects within reinforcement learning procedures. Experience replay, randomized batch sampling, and exploration decay strategies were incorporated into DQN training to improve policy generalization across varying settlement conditions and transaction patterns. While the framework demonstrated strong anomaly ranking capability, continued evaluation under evolving market conditions remains necessary to assess long-term adaptability to new charge codes, regulatory adjustments, and changing operational behaviors.

**H. Competing models**

ANNA framework performance was assessed against six models selected for their relevance to energy market anomaly detection and recent use in settlement auditing studies. Implementing and evaluating additional algorithms would constitute an expansion of the completed experimental study beyond its original comparative scope. The present analysis is therefore limited to the six implemented methods, and no claims are made regarding performance relative to untested alternatives.

- **Statistical Thresholding (±2 SD):** A conventional method for flagging anomalies by identifying data points beyond two standard deviations from the mean, assuming a normal distribution, widely used in ISO settlement processes for basic compliance checks due to its simplicity and low computational cost [2].
- **K-Nearest Neighbors (KNN):** KNN identifies anomalies by measuring the distance of a data point from its nearest neighbors, with greater distances indicating outliers. It has been applied in energy market auditing for its ability to detect local anomalies in dense datasets [10].
- **K-Means Clustering:** This unsupervised method partitions the data into $k$ clusters and detects anomalies as points that lie far from cluster centroids. K-Means is commonly used in energy applications for high-volume settlement grouping and pattern discovery [12].
- **Q-Learning:** Q-Learning is a model-free reinforcement learning algorithm that learns an optimal policy through action-value iteration. It has been proposed for adaptive anomaly detection in dynamic environments such as energy markets [5].
- **Deep Q-Network (DQN):** DQN extends Q-Learning by using a neural network to approximate the Q-function, enabling function approximation beyond a tabular state-action representation. Recent work demonstrates its effectiveness in financial and energy transaction anomaly detection due to its scalability and learning capacity [15].
- **Supervised Multilayer Perceptron (MLP):** A conventional supervised neural-network classifier was evaluated as an additional baseline to assess whether the DQN formulation provided measurable benefit beyond standard supervised neural classification. The MLP was trained using the same standardized settlement Amount input, proxy anomaly labels, preprocessing procedures, and experimental data partitions as the DQN. The MLP consisted of three fully connected hidden layers containing 256, 128, and 64 neurons, respectively, with rectified linear unit (ReLU) activation functions. The output layer was configured for binary classification of normal and anomalous settlement records. The network was trained using the Adam optimizer and binary cross-entropy loss, with predicted probabilities converted to class labels using a decision threshold of 0.50. This configuration provides a supervised neural-network reference for evaluating the DQN formulation under the specified experimental conditions [15]. The MLP was not established as an exhaustively optimized supervised-learning benchmark, and additional class-imbalance treatments and hyperparameter optimization procedures were not documented as part of the completed evaluation. Consequently, the observed performance differences should be interpreted within the limitations of the experimental configuration, proxy-labeling methodology, and class imbalance and do not establish the general superiority of reinforcement learning over supervised classification.

The comparative methods evaluated in this study were selected to reflect techniques commonly employed or representative of current practice within power utility settlement auditing, as well as recent approaches proposed in the literature to improve anomaly detection in complex market environments. Statistical thresholding remains the primary operational method for identifying settlement discrepancies, while K-Nearest Neighbors, K-Means clustering, Supervised MLP, and Q-Learning represent established machine learning and reinforcement learning techniques applicable to structured settlement data. The objective of this evaluation was to assess whether the proposed ANNA framework provides measurable improvement over methods representative of current industry practice rather than to perform an exhaustive comparison against every anomaly detection algorithm reported in the literature. The experimental comparison was limited to the six methods implemented within the original study design. Additional algorithms, including Isolation Forest, gradient boosting, One-Class Support Vector Machines, and autoencoder-based approaches, were not evaluated. These methods may provide useful additional comparisons in future research; accordingly, the present findings establish relative performance only among the evaluated methods and do not demonstrate superiority over untested alternatives.

To promote a fair comparison across all evaluated methods, hyperparameters for the comparative models were selected using established practices from the literature together with validation-based empirical evaluation. For the K-Nearest Neighbors (KNN) classifier, the neighborhood size ($k$) was selected through preliminary validation experiments to provide stable classification performance while minimizing sensitivity to local noise. For K-Means clustering, the number of clusters was determined through exploratory validation analyses and selected to provide stable partitioning of the settlement data. The statistical thresholding baseline employed the conventional ±2 standard deviation criterion, reflecting common operational practice in settlement auditing. The DQN hyperparameters, including learning rate, discount factor, replay buffer size, batch size, target-network update frequency, and exploration schedule, were similarly refined through iterative validation experiments. Although an exhaustive hyperparameter search was beyond the scope of this study because of the computational demands associated with the large-scale dataset, consistent validation procedures were applied to all models to promote an equitable performance comparison.

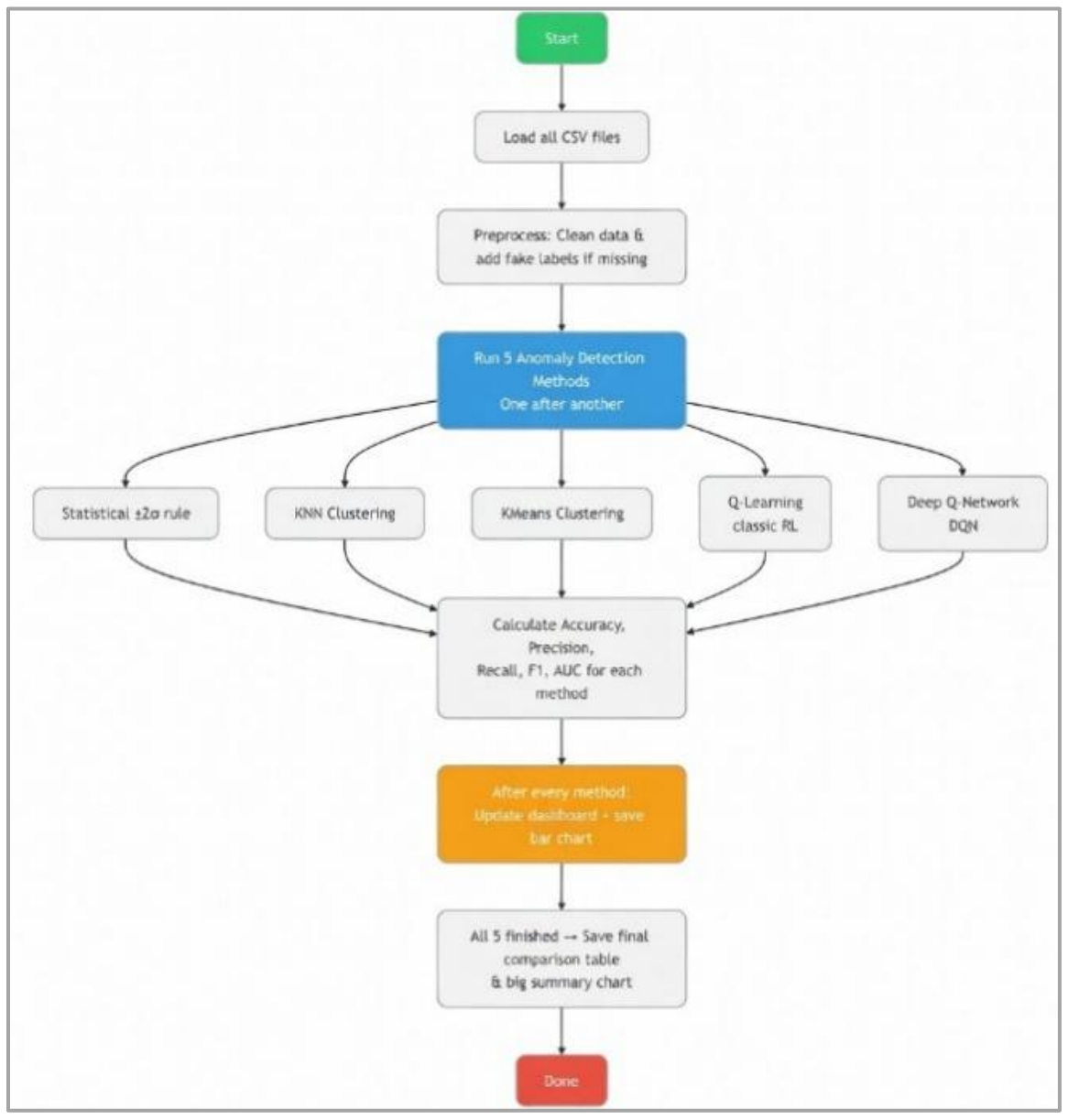

## IV. EXPERIMENT RESULTS

Evaluation of the ANNA framework consists of benchmarking predictive performance and computational efficiency using model-specific experimental subsets derived from the 2.9 million-record labeled working dataset, which was constructed from the broader source dataset of 63 million CAISO settlement records. The results reported in this section therefore correspond to the respective model-specific experimental configurations rather than direct training and evaluation of each model on the complete source dataset. The key classification metrics to identify detection fidelity include precision, recall, F1-score, and accuracy. The area under the Receiver Operating Characteristic (AUC-ROC) is the metric utilized to identify the model's ability to distinguish between normal and anomalous cases across various classification thresholds. Following model detection capabilities, evaluation incorporates system-level metrics regarding total execution time and memory consumption to assess the scalability and deployability of each method under realistic operating conditions. Through this multi-faceted analysis framework, performance conclusions are grounded in predictive power and deployment feasibility in high-volume, time-sensitive energy settlement environments. Because anomaly labels were partially derived from operational heuristics and statistical benchmarking procedures, evaluation results should be interpreted within the context of approximate anomaly classification rather than fully verified forensic audit outcomes.

Stable convergence behavior was generally observed across grouped settlement environments following exploration decay and replay stabilization procedures, although localized performance variability remained present across charge code categories and transaction groupings. Variations in settlement structure, transaction density, and anomaly frequency contributed to differing convergence rates across localized models, reinforcing the importance of grouped evaluation within heterogeneous market environments.

### A. Detection Precision, Recall, and F1-Score

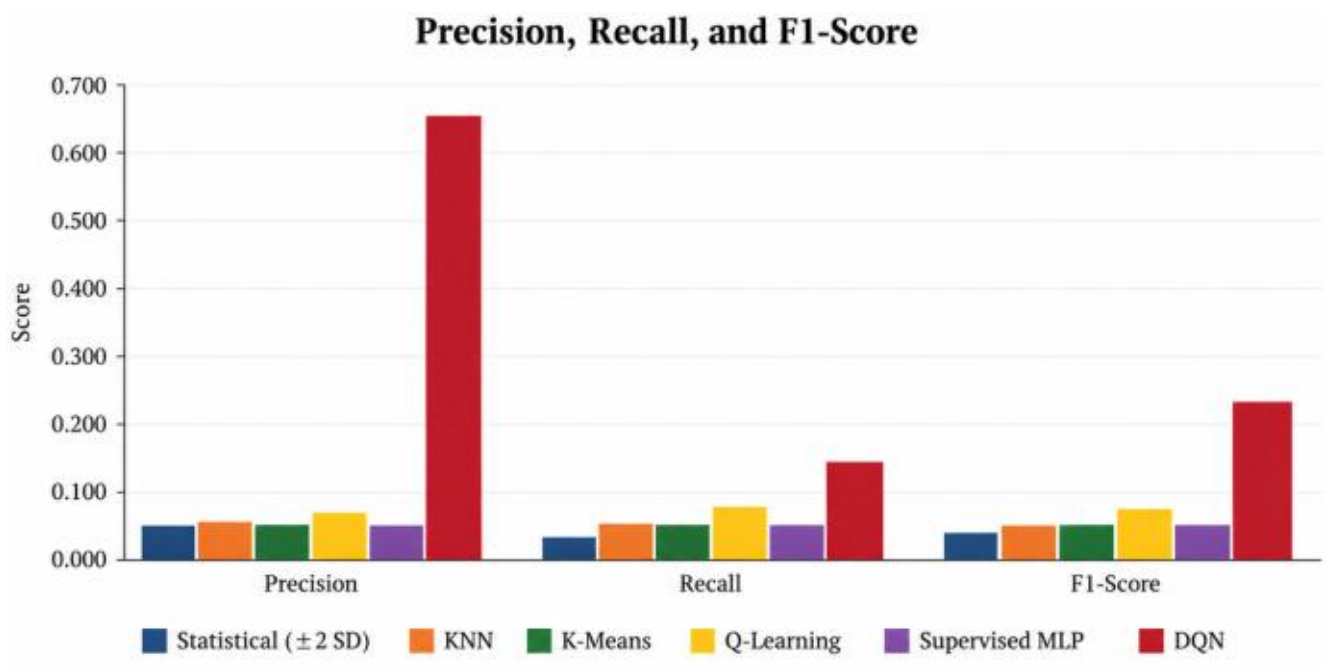

*Fig. 2: Precision, Recall, F1-Score.*

Fig. 2 is a comparative visualization of the precision, recall, and F1-scores for six anomaly detection methods implemented in the ANNA framework: Statistical (±2 SD), KNN, KMeans, Standard MLP, Q-Learning, and Deep Q-Network (DQN).

The Q-Learning model demonstrates modest anomaly detection capability, with average performance metrics of 0.8935 accuracy, 0.0546 precision, 0.0672 recall, and an F1-score of 0.0599 across nine evaluation runs. These results indicate that while the model correctly classifies the majority of non-anomalous transactions, reflected in its relatively high accuracy, it struggles to identify true anomalies. The low precision and recall values show Q-Learning frequently misclassifies anomalous records and fails to detect the majority of them, capturing only about 6.7% of actual anomalies.

With a more realistic detection profile, the DQN model shows a precision of 0.6515, a recall of 0.1415, and an F1-score of 0.2326. While DQN achieves the highest precision among realistic models, the low recall of 0.1415 indicates that approximately 86% of actual anomalies are missed. Recent studies in smart-meter anomaly detection using deep reinforcement learning demonstrate similar challenges and support the need for improved anomaly detection strategies [36]. For a financial auditing system, this represents a significant operational risk, as undetected anomalies may result in substantial financial losses and compliance exposure. Potential contributing factors include class imbalance in the training data, conservative reward function calibration, and the inherent difficulty of detecting subtle anomalies that deviate only marginally from normal transaction patterns.

The Statistical method (±2 SD) exhibited notably poor performance with a precision of 0.0500, a recall of 0.0146, and an F1-score of 0.0226. These scores reflect the inherent rigidity and limited capacity of traditional statistical methods when applied to high-variance or evolving datasets.

The clustering techniques of KNN and KMeans resulted in nearly identical F1-scores for performance of 0.0489 and 0.0516, respectively. The supervised MLP baseline achieved an accuracy of 0.9024, precision of 0.0480, recall of 0.0458, and F1-score of 0.0469. These results indicate limited ability to identify the minority anomaly class despite relatively high overall accuracy. In comparison, DQN achieved substantially higher precision (0.6515), recall (0.1415), and F1-score (0.2326), indicating stronger anomaly detection performance under the evaluated configuration.

Among the evaluated methods, DQN achieved the highest precision and F1-score under the reported experimental configuration; however, its low recall limits conclusions regarding comprehensive anomaly detection. The Q-Learning results are numerically higher than the DQN but lack generalization. Underperformance of KNN, KMeans, and Statistical models reinforces the need for adaptive, learning-based systems in a modern auditing environment, where anomaly detection must operate in noisy, high-dimensional, evolving data landscapes.

**B. Accuracy Comparison Across Models**

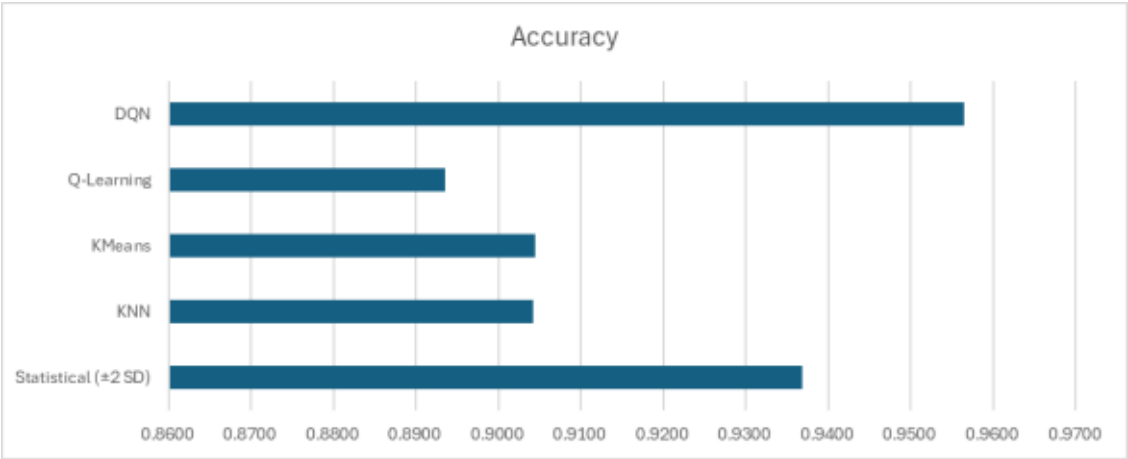

*Fig. 3: Model Accuracy.*

The comparative accuracy achieved by the six anomaly detection models evaluated within the ANNA framework are presented in Fig 3. Accuracy measurement of the proportion of correctly classified observations is a common metric, but its relevance is diminished in terms of anomaly detection characterized by severe class imbalance. In such contexts, high accuracy may improperly indicate a model's tendency to favor majority (normal) classes, thereby obscuring insight into the true detection efficacy of the models.

The DQN model leads with the highest accuracy of 0.9565, reflecting its strong predictive capability and superior ability to balance sensitivity and specificity. Among the evaluated methods, DQN achieved the highest precision and F1-score under the reported experimental configuration; however, its recall remained low, indicating that most proxy-labeled anomalies were not detected. Its ROC-AUC of 0.6829 indicates moderate discriminatory performance. Accordingly, the DQN is more appropriately interpreted as an anomaly-prioritization mechanism than as a comprehensive anomaly detector.

KMeans and KNN models recorded lower accuracies of 0.9045 and 0.9042, respectively. These values are reflective of higher misclassification rates and reinforce the limitation of clustering-based anomaly detection in structured, high-dimensional transaction data. With their unsupervised design and dependence on spatial similarity metrics, both algorithms tend to underperform in environments with overlapping class boundaries and significant noise. Their effectiveness is further constrained by their inability to exploit sequential or contextual information.

The Statistical (±2 SD) method had an intermediate accuracy of 0.9368. While still trailing the DQN model, this performance is tempered by the method's poor detection metrics and rigidity. The method's rule-based nature enables ease of interpretation at the expense of adaptive capabilities required to manage complex, shifting patterns in energy settlement data.

With an accuracy of 0.8935, the Q-Learning algorithm appears to perform well at first glance; however, this value is largely driven by the overwhelming proportion of non-anomalous transactions in the dataset rather than meaningful anomaly detection capability. When considered alongside the model's low precision, recall, and F1-score, the accuracy metric reflects a tendency to default to majority-class predictions rather than learning a generalizable anomaly-detection policy. This behavior is consistent with the limitations of tabular Q-Learning in high-dimensional settlement data, where the agent may converge to simplistic decision rules instead of capturing the underlying structure of anomalous behavior. As such, accuracy alone provides an incomplete and potentially misleading assessment of model performance. More rigorous validation strategies, such as k-fold cross-validation, temporal holdout evaluation, and out-of-distribution testing, are necessary to determine whether the model can generalize effectively under realistic deployment conditions.

The DQN produced the strongest accuracy among the evaluated learning-based methods; however, accuracy should be interpreted alongside its low recall and moderate ROC-AUC because of the class imbalance present in the dataset. The lower accuracies observed for KNN, K-Means, and Statistical thresholding provide additional context for the comparative performance observed under the present experimental configuration.

**C. Receiver Operating Characteristic (AUC-ROC) Evaluation**

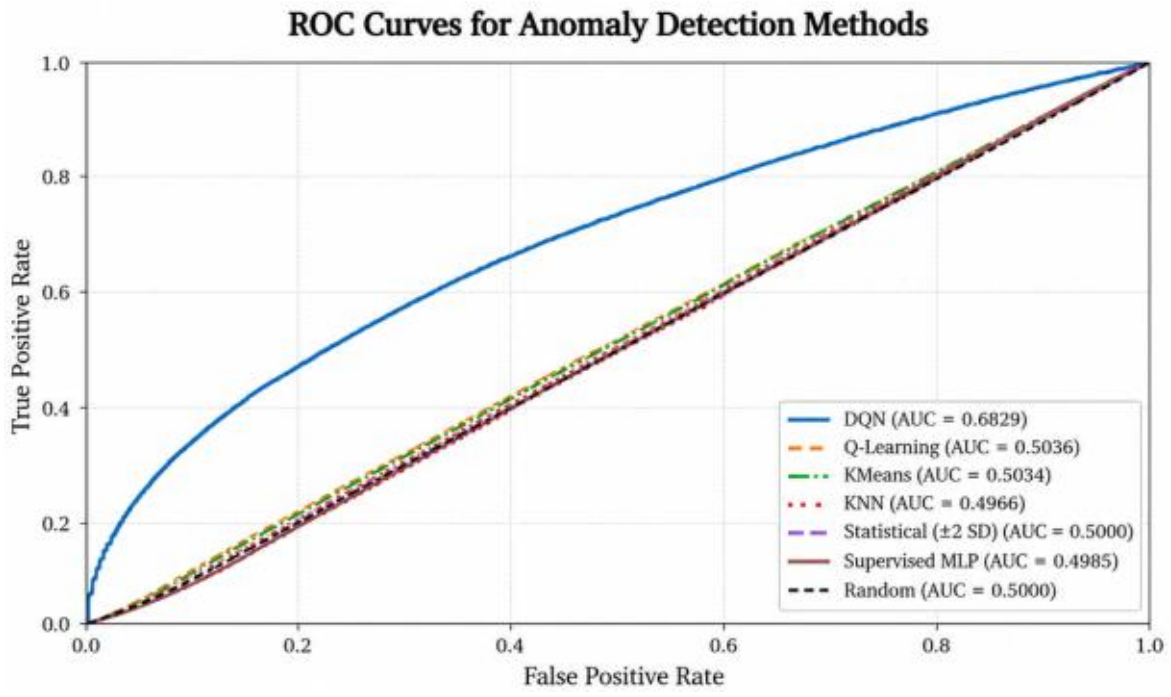

*Fig. 4: Area Under the ROC.*

The Area Under the Receiver Operating Characteristic (AUC-ROC) curve is a single scalar value essential for evaluating the performance of anomaly detection models across a spectrum of classification thresholds by summarizing the trade-off between sensitivity (true positive rate) and specificity (false positive rate). The closer the AUC-ROC is to 1.0, the stronger the discriminatory power of the model. Values near 0.5 imply a lack of effective separation between anomalous and normal instances, akin to random guessing.

Figure 4 shows the DQN substantially outperformed all other evaluation methods with the highest AUC-ROC score of 0.6829. The results indicate the DQN's ability to detect anomalies and rank transactions effectively based on anomalous characteristics. The model's performance strongly aligns with the audit context, where prioritization and rank-based risk scoring are critical to efficient review and triage.

The KMeans and KNN models demonstrated considerably weaker AUC-ROC values of 0.5034 and 0.4966, slightly above and below a random baseline. Their ROC curves closely follow the diagonal midline, indicating minimal discriminatory capability across thresholds. The supervised MLP produced an AUC-ROC of 0.4985, indicating approximately chance-level discrimination between normal and proxy-labeled anomalous transactions. In comparison, the DQN achieved an AUC-ROC of 0.6829, demonstrating moderate discriminatory capability. The supervised comparison therefore provides evidence that the DQN's observed performance was not attributable solely to the use of a neural-network function approximator. However, this result is specific to the evaluated dataset and experimental configuration and does not establish a general superiority of reinforcement learning over supervised neural classification.

Q-Learning exhibited a similar limitation, producing an AUC-ROC of 0.5036 despite its superficially strong accuracy. This near-random ranking performance indicates that the model fails to assign meaningful confidence scores to anomalous transactions. The discrepancy between its accuracy and its weak AUC-ROC reflects the model's tendency to default to majority-class predictions rather than learning a robust anomaly-detection policy. As a result, Q-Learning provides limited utility in auditing contexts where prioritization, interpretability, and graded anomaly scoring are essential.

Consistent with its threshold-based binary classification structure, the Statistical (±2 SD) method yielded an AUC-ROC of 0.5000. This result indicates limited ranking capability for the Statistical (±2 SD) method under the evaluated configuration. Underscoring the inflexibility of traditional methods and the need for adaptive audit systems once more.

DQN demonstrated stronger ranking performance than the evaluated statistical and clustering baselines, supporting its potential use as an anomaly-prioritization mechanism for analyst review.

**D. Precision–Recall Curve and PR-AUC Evaluation**

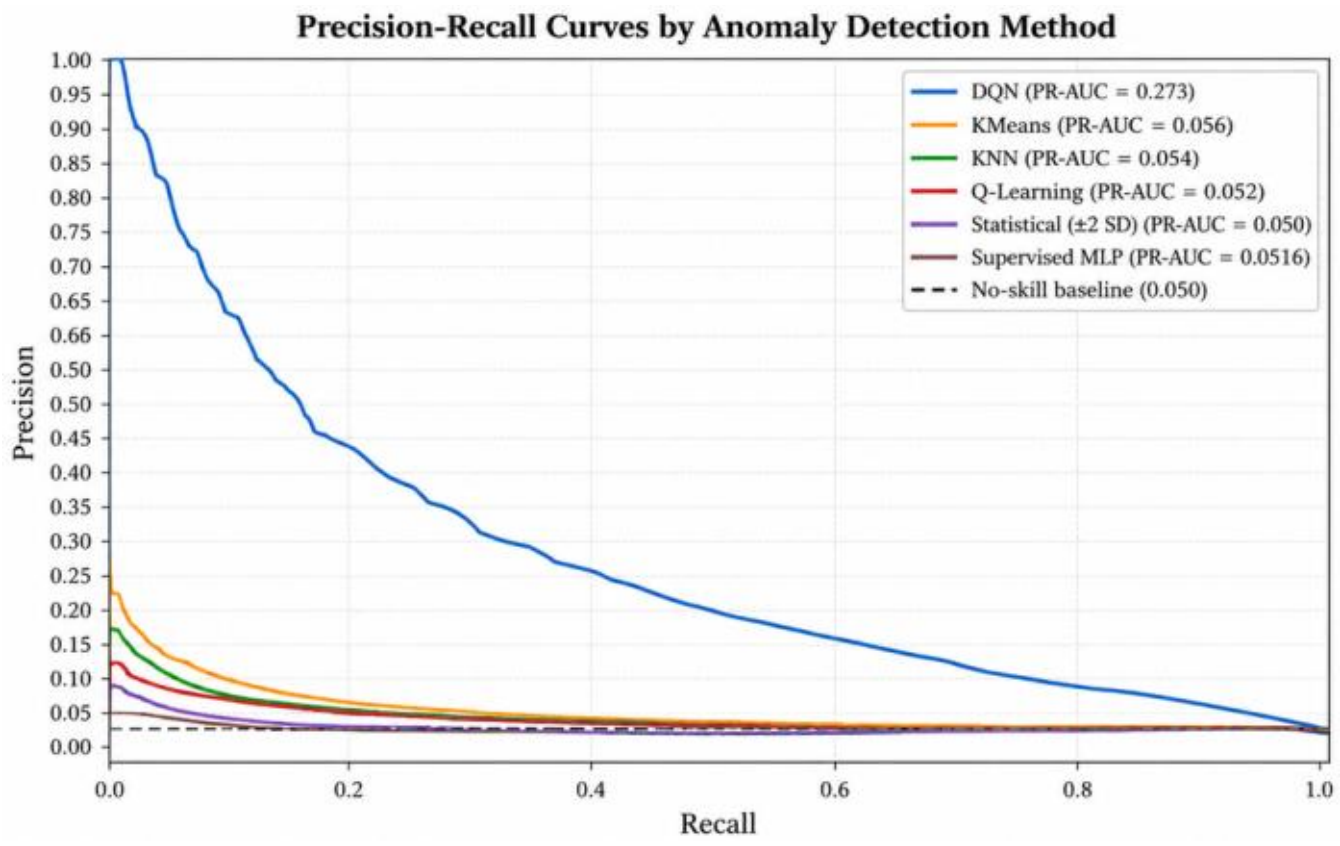

*Fig. 5: PR-AUC*

Precision–Recall analysis was performed to provide an additional threshold-independent evaluation of anomaly detection performance under severe class imbalance. Unlike ROC-AUC, which summarizes the trade-off between true-positive and false-positive rates, PR-AUC emphasizes performance on the anomalous minority class by directly evaluating the relationship between precision and recall across decision thresholds. This metric is particularly relevant in the present study because anomalous settlement transactions account for approximately 4.99% of the evaluated dataset.

Fig. 5 presents the precision–recall curves for the Statistical, KNN, K-Means, Q-Learning, supervised MLP, and DQN methods. The corresponding PR-AUC values were 0.273 for DQN, 0.056 for K-Means, 0.054 for KNN, 0.052 for Q-Learning, 0.050 for Statistical (±2 SD), and 0.0516 for the supervised MLP. The no-skill reference level corresponds approximately to the anomaly prevalence in the evaluation data. The DQN curve remains substantially above the other methods across a broad range of recall values, indicating stronger anomaly-ranking capability and improved concentration of true anomalies among the highest-scored transactions. This result is consistent with the DQN's reported AUC-ROC of 0.6829 and its comparatively high operating-point precision of 0.6515.

KNN, K-Means, Q-Learning, and Statistical thresholding remain relatively close to the no-skill baseline for much of the recall range, indicating limited discrimination of anomalous transactions. Among these methods, K-Means and KNN provide modest improvement at low recall, while Q-Learning and Statistical thresholding demonstrate little threshold-independent ranking capability. These findings reinforce the limitations of fixed statistical and conventional clustering approaches in highly imbalanced, heterogeneous settlement data. The supervised MLP achieved a PR-AUC of 0.0516, remaining close to the approximately 4.99% no-skill baseline associated with anomaly prevalence. This result is consistent with its ROC-AUC of 0.4985 and indicates limited ability to rank anomalous transactions under the evaluated supervised-learning configuration. In contrast, DQN demonstrated substantially stronger threshold-independent anomaly discrimination.

The PR-AUC analysis provides important context for interpreting the fixed-threshold precision, recall, and F1-score results. Although DQN achieves substantially higher precision than the comparison methods, its recall of 0.1415 indicates that performance remains concentrated within a relatively narrow subset of high-confidence anomalies. Therefore, DQN is most appropriately interpreted as an anomaly-prioritization and analyst-support mechanism rather than a comprehensive autonomous detector. Further improvements should focus on reward redesign, class-imbalance mitigation, threshold calibration, and ensemble modeling to increase recall without substantially reducing precision.

**E. Memory Utilization and Computational Efficiency**

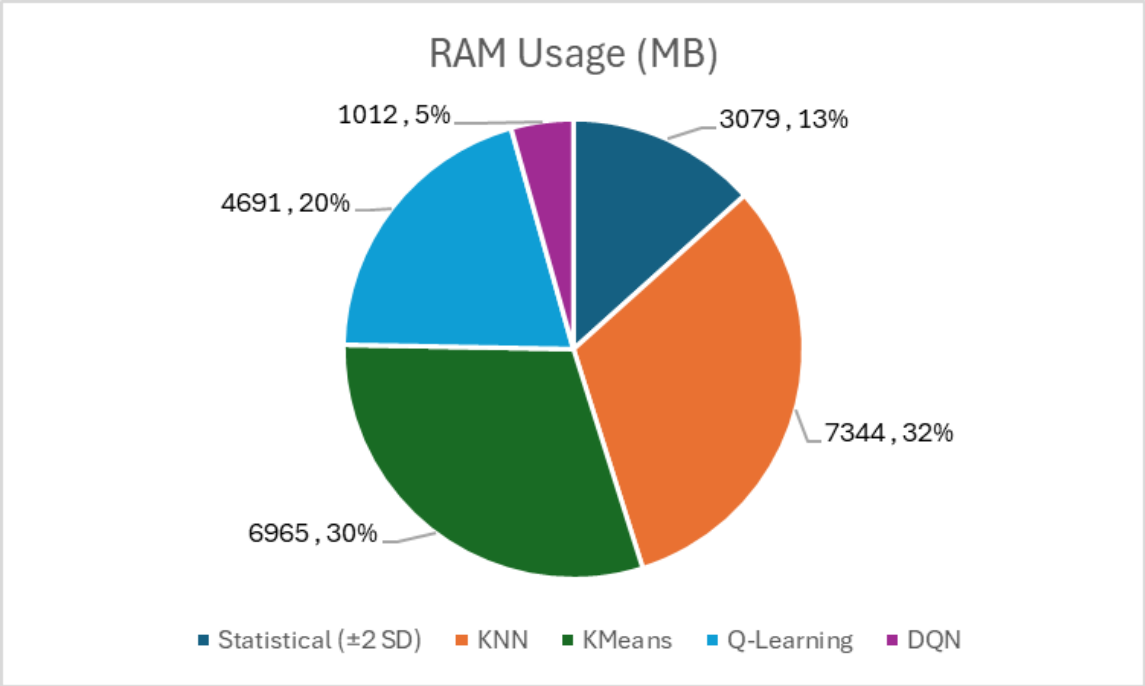

*Fig. 6: RAM Usage (MB).*

A critical benchmark for evaluating the feasibility of anomaly detection models is memory consumption, particularly in high-throughput environments within real-time or large-scale energy settlement auditing systems. Figure 6 presents the comparative RAM usage observed across the five methods integrated and evaluated within the ANNA framework.

DQN demonstrated the most efficient memory utilization, requiring only 1011.87 MB of RAM. In contrast, the KNN and KMeans models were the most memory-intensive, consuming 7343.68 MB and 6965.38 MB, respectively.

The Statistical (±2 SD) method and Q-Learning algorithm exhibited moderate memory consumption at 3079.12 MB and 4691.09 MB. Memory utilization of the statistical approach is due to internal caching mechanisms for rolling averages and variance calculations. For Q-Learning, memory usage is driven by the linearly scaling tabular Q-Value storage structure with observed states and actions.

The observed memory utilization indicates that the DQN required less memory than the evaluated comparison methods under the present experimental configuration. However, broader deployment efficiency would require validation under larger and more representative operational workloads.

**F. Runtime Performance**

| Method | Total Runtime (s) | Training (s) | Inference (s) |
|---|---|---|---|
| Statistical (±2 SD) | 5.9446 | 0.0017 | 0.0041 |
| KNN | 7.4556 | 0.0341 | 0.1984 |
| K-Means | 7.1262 | 0.4683 | 0.0020 |
| MLP | 11.3437 | 2.3659 | 0.0060 |
| Q-Learning | 17.7024 | 10.6401 | 0.0030 |
| DQN | 63.2662 | 56.5177 | 0.0040 |

*Fig. 7: Results Summary.*

Runtime efficiency is an important consideration when evaluating anomaly-detection methods for operational energy-settlement environments, particularly where large transaction volumes and timely analytical responses are required. To improve the comparability and interpretability of the computational evaluation, all six methods were benchmarked using a common 10,000-record workload selected chronologically by Trade Date from the 2.9 million-record labeled working dataset. Runtime was further decomposed into model training and inference stages to distinguish the computational cost of model development from post-training transaction evaluation. Figure 7 presents the resulting runtime performance across the evaluated methods.

Total execution time varied considerably among the models. The Statistical (±2 SD) method required 5.9446 seconds, while K-Means and KNN required 7.1262 and 7.4556 seconds, respectively. The supervised MLP required 11.3437 seconds, Q-Learning required 17.7024 seconds, and DQN had the highest total runtime at 63.2662 seconds. For the MLP, Q-Learning, and DQN models, the common 10,000-record workload was chronologically partitioned into 6,000 training, 1,500 validation, and 2,500 test records. The Statistical, KNN, and K-Means methods operated on the complete 10,000-record workload without a held-out training-validation split, consistent with their respective implementations.

The higher total runtime of DQN was primarily attributable to model training. DQN required 56.5177 seconds for training, compared with 10.6401 seconds for Q-Learning and 2.3659 seconds for the supervised MLP. In contrast, post-training inference represented a substantially smaller computational component. DQN required 0.0040 seconds to perform inference on the 2,500-record test subset. The measured inference times for the Statistical, KNN, and K-Means methods were 0.0041, 0.1984, and 0.0020 seconds, respectively. The corresponding MLP and Q-Learning inference measurements were approximately 0.0060 and 0.0030 seconds under the computational benchmark.

## V. DISCUSSION AND FUTURE WORK

From an institutional perspective, the increasing complexity of ISO settlement structures has significantly expanded the operational burden associated with manual auditing and exception review. As transaction volumes continue to increase across interconnected market systems, scalable anomaly prioritization frameworks may help reduce review latency, improve settlement transparency, and support more efficient allocation of audit resources. In this context, reinforcement learning approaches offer potential value not only through predictive capability but also through their ability to adapt alongside evolving market structures and operational behaviors.

### A. Detection Precision, Recall, and F1-Score

Building on the empirical results presented above, several observations warrant deeper interpretation. While high performance on historical datasets is encouraging, such outcomes often fail to generalize under dynamic or previously unseen market conditions. This reinforces the critical role of robust evaluation strategies, such as cross-validation and temporal holdout testing, to assess model resilience and adaptability beyond static snapshots of data.

Among all evaluated models, the Deep Q-Network (DQN) demonstrated superior balance between sensitivity and specificity, as evidenced by its consistently higher F1-score. This metric, which harmonizes precision and recall, is especially important in high-stakes financial auditing where false positives can overwhelm analysts and degrade confidence in automated alerts. DQN's comparatively higher precision and F1-score indicate potential value for prioritizing high-risk transactions for analyst review; however, its low recall limits conclusions regarding comprehensive anomaly detection.

Conversely, fixed-threshold statistical models struggled to detect subtle or temporally dispersed anomalies often found in complex settlement data. Their static boundaries fail to adapt to shifting distributions or context-specific variations. Similarly, unsupervised clustering methods using low-dimensional input features performed only marginally better than the baseline. These approaches are often unable to resolve overlapping or irregular anomaly distributions, especially when sparse outliers are aggregated into loosely defined clusters.

### B. Accuracy Comparison Across Models

Although accuracy is a standard classification metric, its interpretive value diminishes in imbalanced datasets such as energy settlement records, where anomalies constitute a small minority. High overall accuracy in this context may reflect a model's bias toward dominant (non-anomalous) classes rather than genuine detection capability. Thus, precision, recall, and F1-score are more appropriate for evaluating performance in rare-event detection scenarios.

### C. Receiver Operating Characteristic (AUC-ROC) Evaluation

The Area Under the ROC Curve (AUC-ROC) offers a threshold-independent assessment of model performance across varying decision criteria. The clustering-based methods remained near chance-level discrimination, whereas DQN achieved moderate discriminatory performance under the evaluated configuration. This limitation further emphasizes the advantages of reinforcement learning approaches, which dynamically optimize policies based on continuous feedback and evolving data patterns.

### D. Memory Utilization and Computational Efficiency

Clustering algorithms incurred significant computational overhead, primarily due to their reliance on pairwise distance calculations, iterative fitting, and index storage. While simple in concept, these models scale poorly as data volume and dimensionality increase, making them impractical for real-time or large-scale deployment. In contrast, the DQN architecture, though computationally intensive during training, supports inference efficiency and modular scalability, positioning it as a more viable solution for production-level auditing systems.

**E. Key Takeaways**

The experimental results demonstrate the performance of the ANNA (Automated Neural Network Auditing) framework when benchmarked against traditional and unsupervised anomaly detection methodologies. Across the six evaluated methods, DQN achieved the strongest anomaly-discrimination performance under the reported experimental configuration, although it incurred the highest total computational cost and exhibited low recall. Importantly, these evaluations were conducted using grouped and filtered subsets derived from a broader settlement dataset exceeding 63 million records, ensuring experimental analysis remained computationally feasible while still reflecting realistic market behaviors and transaction variability.

The traditional statistical method (±2 SD) set the baseline for performance but provided minimal utility in anomaly detection tasks. While conceptually simple, accessible, and computationally inexpensive, the fixed-threshold logic of the method resulted in limited precision of 0.0500, recall of 0.0146, and corresponding F1-score of 0.0226, indicating an inability to identify subtle or emerging anomalies in complex, temporally irregular datasets. Combined with this model's relatively high runtime (5.9446 s), its operation values for real-time or high-frequency auditing applications are greatly diminished.

The unsupervised KNN and KMeans clustering methods offered slight improvements in detection performance at the expense of computational inefficiency. Each model achieved F1-scores of 0.0489 and 0.0516, indicating marginal gains over the statistical method. Their moderate precision and recall scores suggest the models struggle with high-dimensional overlap and noisy market conditions. Combined with the highest RAM usage amount of all methods (7343.68 MB for KNN and 6965.68 MB for KMeans) illustrates the clustering methods are ill-suited for deployment in memory-constrained or real-time auditing systems. The supervised MLP comparison further supports this distinction: despite using the same underlying anomaly-labeling framework and comparable input information, the MLP achieved near-random discrimination (AUC-ROC = 0.4985), whereas DQN achieved moderate discrimination (AUC-ROC = 0.6829).

In contrast, the Q-Learning architecture produced an accuracy of 0.8935, but this value is misleading when considered alongside its low precision (0.0546), recall (0.0672), and F1-score (0.0599). These metrics indicate the agent predominantly predicts the majority class and fails to identify true anomalies. The near-random AUC-ROC of 0.5036 further demonstrates the model does not learn a meaningful decision boundary. This behavior is consistent with the limitations of tabular Q-Learning in high-dimensional settlement data, where the algorithm tends to memorize frequent patterns rather than develop a generalizable anomaly-detection policy.

Ultimately, DQN demonstrated the strongest anomaly-discrimination performance among the evaluated methods under the reported experimental configuration. In practical auditing environments, anomaly detection effectiveness depends not only on predictive capability but also on the ability to support analyst review and regulatory defensibility. Within the ANNA framework, flagged transactions retain supporting contextual information including deviation benchmarks, transaction metadata, confidence scoring, and reinforcement learning policy outputs to assist downstream audit interpretation. This layered approach enables auditors to prioritize high-risk settlements while preserving traceability necessary for compliance review, settlement dispute analysis, and post-audit reconciliation procedures.

The DQN precision of 0.651, recall of 0.1415, and F1-score of 0.2326 are significant improvements over the performance observed in the traditional and clustering methods. The use of temporally separated validation procedures and grouped evaluation subsets provided additional confidence that the observed performance improvements were not solely the result of memorized settlement behavior within static datasets. Its AUC-ROC of 0.6829 demonstrates superior ranking capability to distinguish anomalous transactions across various decision thresholds. DQN had the lowest reported memory requirement (1011.9 MB), while post-training inference required 0.0040 seconds on the 2,500-record test subset. However, the DQN recall of 0.1415 remains a critical limitation for operational deployment, as missing approximately 86% of true anomalies poses substantial financial and compliance risks. Future work should pursue concrete strategies to improve recall, including cost-sensitive reward redesign with increased false-negative penalties, synthetic minority oversampling (SMOTE) to address class imbalance, ensemble approaches combining DQN with complementary detection methods, and threshold tuning to optimize the precision-recall trade-off for the specific risk tolerance of the auditing context.

To assess the stability of the proposed DQN framework, performance metrics were summarized across eleven experimental evaluations. Table 4 reports the mean and standard deviation for the principal evaluation metrics. Although the experiments were conducted under a consistent experimental configuration, the observed variability provides an indication of the model's sensitivity to differences in the evaluated settlement groups and demonstrates the reported performance is representative of multiple evaluations rather than a single isolated result.

**Deep Q-Network (DQN) Performance Across Experimental Evaluations (Mean ± SD)**

| Performance Metric | Mean | Standard Deviation | Mean ± SD |
|---|---|---|---|
| Precision | 0.652 | 0.229 | **0.652 ± 0.229** |
| Recall | 0.142 | 0.143 | **0.142 ± 0.143** |
| F1-Score | 0.203 | 0.189 | **0.203 ± 0.189** |
| ROC-AUC | 0.683 | 0.128 | **0.683 ± 0.128** |

*Table 4: DQN Performance*

As shown in Table 4, the DQN achieved a mean ROC-AUC of 0.683 ± 0.128, while maintaining a mean precision of 0.652 ± 0.229. The comparatively larger variation observed in precision, recall, and F1-score reflects the heterogeneous characteristics and class imbalance present across the evaluated settlement groups. These results indicate the proposed framework produces consistent discriminatory performance while exhibiting expected variability in threshold-dependent classification metrics when applied to diverse settlement scenarios.

### F. Internal and External Threats to Study Results

The ANNA detection engine demonstrates accuracy and adaptability in detecting anomalous settlements in the real-world energy market. Still, it is subject to practical and systemic limitations ranging from technical, organizational, and regulatory domains. The limitations in each domain affect scalability, reliability, and adoption of AI-based auditing. Obstacles encountered during the design and deployment of the ANNA, and broader limitations inherent to applying deep learning and reinforcement learning frameworks to high-integrity financial environments include:

- **Computation Complexity of Deep Learning Training:** The integration of DQN reinforcement learning with large-scale settlement data introduces additional computational requirements associated with iterative policy updates and neural-network training. Under the standardized computational benchmark used in this study, DQN incurred the highest total computational cost among the evaluated methods, with most of this cost attributable to model training. Specifically, DQN required 63.2662 seconds of total execution time, including 56.5177 seconds for training, while post-training inference required 0.0040 seconds on the 2,500-record test subset. This distinction indicates low post-training inference latency under the evaluated configuration but does not, by itself, establish real-time operational readiness. Scaling the framework to larger datasets, additional settlement features, or multiple model instances may increase computational requirements and could motivate the use of parallel processing, model optimization, or distributed training strategies. Furthermore, changes in charge codes, market behavior, or regulatory requirements may require periodic model retraining and revalidation. Future iterations of the framework may therefore investigate incremental learning or rolling retraining procedures to improve responsiveness to long-term changes in settlement data distributions. Although reinforcement learning provides a mechanism for incorporating asymmetric decision consequences through reward design, its performance remains sensitive to reward specification, exploration behavior, and changes in the underlying data distribution. As settlement markets evolve through regulatory updates, changing participation patterns, and emerging transaction structures, maintaining model stability and generalization will require continued validation and methodological refinement.

- **Data Availability and Privacy Concerns:** Anomaly detection relies on high-resolution, granular transaction settlement data with accurate labelling, which is not always available across system operation or for all market participants. Transaction data often contains sensitive personally identifiable information, representing a risk to data sharing, model auditability, and compliance with data governance standards. Future considerations for ANNA may include federated learning or differential privacy techniques to address availability challenges without compromising data integrity or security.
- **Ground-Truth Label Uncertainty:** Large-scale energy settlement datasets rarely contain comprehensively verified anomaly labels due to the operational burden associated with manual forensic review. The labeling methodology utilized in this study combined statistical thresholds with domain-informed heuristics to approximate anomalous settlement behavior. Although representative subsets were manually reviewed to improve confidence in the resulting classifications, some degree of label uncertainty and heuristic bias may remain. Future work should incorporate expanded expert review procedures, semi-supervised learning approaches, and independently validated anomaly cases to strengthen evaluation reliability and reduce dependence on rule-derived labeling strategies.
- **Regulatory Interpretability Constraints:** Regulatory auditing with AI-based systems introduces concerns around transparency, explainability, and compliance. Regulators may require interpretable decision-making processes to validate or contest audit outcomes. ANNA's retention of predicted values and deviation scores for each identified anomaly may still be considered a "black box" methodology insufficient to satisfy regulatory requirements supporting a valid audit claim. Standardization and familiarity with AI-integrated audit trails will play a pivotal role in establishing and maintaining trust and regulatory acceptance of AI-based systems. To address this limitation, future development of ANNA should incorporate model-agnostic interpretability techniques such as LIME (Local Interpretable Model-agnostic Explanations) and SHAP (SHapley Additive exPlanations) to generate human-readable justifications for each flagged anomaly, enabling auditors to evaluate and defend detection outcomes in regulatory proceedings. Future work may also incorporate privacy-preserving RL and explainable anomaly detection frameworks to meet regulatory requirements [32–33].
- **Baseline Comparisons and Generalizability:** ANNA was evaluated against representative statistical, clustering, supervised neural-network, and reinforcement learning approaches commonly used in anomaly detection for energy settlement auditing, including a supervised multilayer perceptron (MLP) baseline. However, the experimental comparison did not include several other contemporary machine learning methods, such as Isolation Forest, One-Class Support Vector Machines, and Autoencoder-based anomaly detection [2], [9], [12], [18]. Consequently, the reported performance should be interpreted as demonstrating effectiveness relative to the selected baseline methods rather than establishing superiority over all available anomaly detection techniques. Future work will incorporate these additional baseline models and broader benchmark datasets to further evaluate the generalizability and comparative performance of the proposed framework.
- **Label Validation:** The anomaly labels used in this study were derived from statistical thresholds and domain-informed business rules and therefore represent proxy labels rather than independently verified ground truth [3], [12]. While representative samples were manually reviewed to improve confidence in the labeling process, comprehensive expert validation and formal inter-rater agreement analysis were beyond the scope of the present study. As a result, the reported performance metrics should be interpreted with this limitation in mind. Future work will incorporate larger expert-validated datasets to further strengthen model evaluation. The present DQN implementation should therefore be viewed as an adaptive decision-optimization framework operating on proxy-labeled data rather than as a demonstration of complex sequential control. Future work will extend this comparison to additional supervised architectures and independently validated datasets to further assess the conditions under which reward-based reinforcement learning provides practical advantages.
- **Reproducibility and Confidentiality:** The study reports the model architectures, available experimental configurations, evaluation metrics, and computational environment used in the completed evaluation. However, the underlying settlement records and certain proprietary data-processing and operational review procedures are subject to confidentiality and data-use restrictions. Consequently, complete public replication using the original dataset and disclosure of restricted operational details are not possible. Additional experimental details and variability estimates not established during the completed evaluation are not reported retrospectively. These limitations constrain independent reproducibility and the interpretation of the findings.

Although explainability mechanisms can improve transparency, anomaly classifications generated within complex reinforcement learning systems should continue to function as analyst-support indicators rather than fully autonomous audit determinations, particularly in high-risk regulatory or financial dispute environments.

## VI. CONCLUSIONS

With the energy sector's continued digital transformation, increased complexity and volume of financial settlements necessitate a transition from traditional auditing methods to more advanced tools capable of efficiently identifying and prioritizing potentially anomalous behavior. This paper outlines the design, implementation, and evaluation of a hybrid AI-based Automated Neural Network Auditing (ANNA) auditing system, integrating statistical benchmarking, unsupervised clustering, and reinforcement learning for transaction settlement anomaly detection. Through modular architecture and data-driven methodology, the ANNA framework addresses critical needs for scalability, accuracy, and interpretable auditing solutions in dynamic energy markets.

### A. Summary of Contributions

- The application of Deep Q-Networks (DQNs) in a simulated Gymnasium-based environment enabled the detection of anomalous settlement transactions through reward-optimized policy learning.
- Application of statistical deviation metrics and unsupervised clustering to benchmarks and characterize traditional transactional outliers across multiple Positions and Settlement Types to improve system sensitivity to diverse patterns of anomalous behaviors.
- Reinforcement learning with asymmetric reward shaping to optimize anomaly classification by assigning different consequences to false-positive and false-negative decisions in support of actionable audit outcomes.
- Utilization of model-specific training and evaluation subsets derived from a 2.9 million-record labeled working dataset, which was constructed from a broader source dataset of 63 million CAISO settlement records spanning approximately three years.
- A supervised MLP baseline was additionally evaluated to directly compare reinforcement learning with conventional neural classification; under the evaluated configuration, DQN demonstrated stronger anomaly discrimination than the MLP baseline.

The ANNA framework demonstrated potential for identifying and prioritizing anomalous settlement transactions beyond those identified by the evaluated statistical and clustering baselines. The study additionally highlights the practical challenges associated with large-scale settlement anomaly labeling, emphasizing the need for scalable validation strategies capable of balancing operational feasibility with evaluation reliability.

### B. Real-World Impact and Industry Relevance

The ANNA framework's real-world practical application enhances market settlements by automating the identification of settlement anomalies caused by pricing discrepancies, billing inconsistencies, and potential data manipulation. The ability to adapt to evolving market conditions and transactions is a significant improvement over the existing static compliance framework. Integrating anomaly explanation, forecasted values, and post-detection audit support enhances critical system transparency and traceability for real-world deployment.

### C. Adoption by ISOs and Regulatory Potential

The modular design of ANNA provides a basis for further investigation across ISO settlement environments. However, deployment for regulatory compliance, dispute resolution, or settlement reconciliation would require additional validation, interpretability assessment, and testing under operational conditions.

### D. Conclusion

Although the framework demonstrated encouraging performance improvements relative to traditional statistical and clustering approaches, the current implementation should be viewed primarily as an analyst-support and anomaly prioritization system rather than a fully autonomous auditing replacement.

The results demonstrate the potential of ANNA to support computationally efficient anomaly prioritization in energy-market settlement auditing. However, the DQN's recall of 0.1415 and moderate ROC-AUC of 0.6829 indicate that further validation and methodological improvement are required before broader operational deployment can be established.

The ANNA framework represents a promising advancement in reinforcement learning-assisted settlement auditing for modern energy markets. The framework demonstrates the practical potential of adaptive anomaly detection techniques within increasingly complex settlement environments while highlighting areas requiring continued refinement and operational validation.

Results further indicate existing anomaly detection approaches commonly used in practice, including standard deviation thresholds, K-Means, KNN, and conventional Q-Learning perform poorly across several important evaluation metrics, reinforcing the need for more adaptive and scalable auditing methodologies. The ANNA framework demonstrates the potential of reinforcement learning-assisted anomaly prioritization within the evaluated energy-market settlement dataset. Under the reported experimental configuration, DQN achieved a precision of 0.6515, recall of 0.1415, and ROC-AUC of 0.6829. These results indicate useful prioritization capability for a subset of proxy-labeled anomalies, while the low recall and moderate discrimination limit conclusions regarding comprehensive detection or operational readiness. Further validation would be necessary before deployment as an autonomous settlement-auditing system.